\pdfoutput=1
\documentclass[letterpaper, 10 pt, conference]{ieeeconf}  

\IEEEoverridecommandlockouts                              

\usepackage{graphicx}
\usepackage{epsfig} 
\usepackage{amsmath, amssymb, amsfonts} 
\usepackage{mathtools}
\usepackage[dvipsnames,table,xcdraw]{xcolor}
\usepackage{eucal}
\usepackage{indentfirst}
\usepackage{algorithm}
\usepackage{physics}
\usepackage{multirow}
\usepackage{multicol}
\usepackage{url}
\usepackage{cite}
\usepackage{subcaption}
\usepackage[export]{adjustbox}
\usepackage{hyperref}
\usepackage{bm}
\usepackage{algpseudocode}
\hypersetup{hidelinks}

\newcommand{\sh}{\textsc{h}}
\newcommand{\shbar}{\overline{\textsc{h}}}

\begin{document}
\title{\LARGE \bf
Whole-Body Dynamic Telelocomotion: A Step-to-Step Dynamics Approach to Human Walking Reference Generation
}

\author{Guillermo Colin$^{1}$, Joseph Byrnes$^{2}$, Youngwoo Sim$^{1}$, Patrick M.~Wensing$^{3}$, and Joao Ramos$^{1}$ 
\thanks{This work is supported by the National Science Foundation via grant CMMI-2043339.}
\thanks{$^{1}$ Department of Mechanical Science and Engineering at the University of Illinois at Urbana-Champaign, USA.{\tt\small (gjcolin3@illinois.edu)}} 
\thanks{$^{2}$ Department of Industrial and Enterprise Systems Engineering at the University of Illinois at Urbana-Champaign, USA.} 
\thanks{$^{3}$ Department of Aerospace and Mechanical Engineering at the University of Notre Dame, USA.} 
}

\maketitle

\begin{abstract}
Teleoperated humanoid robots hold significant potential as physical avatars for humans in hazardous and inaccessible environments, with the goal of channeling human intelligence and sensorimotor skills through these robotic counterparts. Precise coordination between humans and robots is crucial for accomplishing whole-body behaviors involving locomotion and manipulation. To progress successfully, dynamic synchronization between humans and humanoid robots must be achieved. This work enhances advancements in whole-body dynamic telelocomotion, addressing challenges in robustness. By embedding the hybrid and underactuated nature of bipedal walking into a virtual human walking interface, we achieve dynamically consistent walking gait generation. Additionally, we integrate a reactive robot controller into a whole-body dynamic telelocomotion framework. Thus, allowing the realization of telelocomotion behaviors on the full-body dynamics of a bipedal robot. Real-time telelocomotion simulation experiments validate the effectiveness of our methods, demonstrating that a trained human pilot can dynamically synchronize with a simulated bipedal robot, achieving sustained locomotion, controlling walking speeds within the range of 0.0~m/s to 0.3~m/s, and enabling backward walking for distances of up to 2.0~m. This research contributes to advancing teleoperated humanoid robots and paves the way for future developments in synchronized locomotion between humans and bipedal robots.
\end{abstract}

\section{Introduction}

The full potential of teleoperated humanoid robots has yet to be realized, particularly in their role as physical avatars for humans in hazardous or inaccessible locations \cite{TRO_survey}. Although noteworthy achievements have been made in loco-manipulation tasks, performance remains limited \cite{loco_mani_ex1, loco_mani_ex2}. This is because for loco-manipulation task completion precise coordination of whole-body dynamics is essential \cite{loco_manipulation_ex1, loco_manipulation_ex2}. As a result, in teleoperated humanoid robot research a major obstacle is the lack of synchronized locomotion between human and robot, which entails coordinated movements like simultaneous lifting-off and stepping-down. The disparity in locomotion dynamics between human and robot makes it challenging to fully leverage human capability. Hence, establishing a dynamic link between human and robot locomotion is crucial. In recent years, efforts have been made to realize this critical behavior \cite{SyncWalk_ex1, SyncWalk_ex2, loco_mani_ex2}. However, this area of research remains largely unexplored, particularly within the bipedal locomotion control community.

\begin{figure}[t]
\begin{center}
\includegraphics[width=0.99\linewidth, clip, trim=100mm 80mm 10mm 30mm]{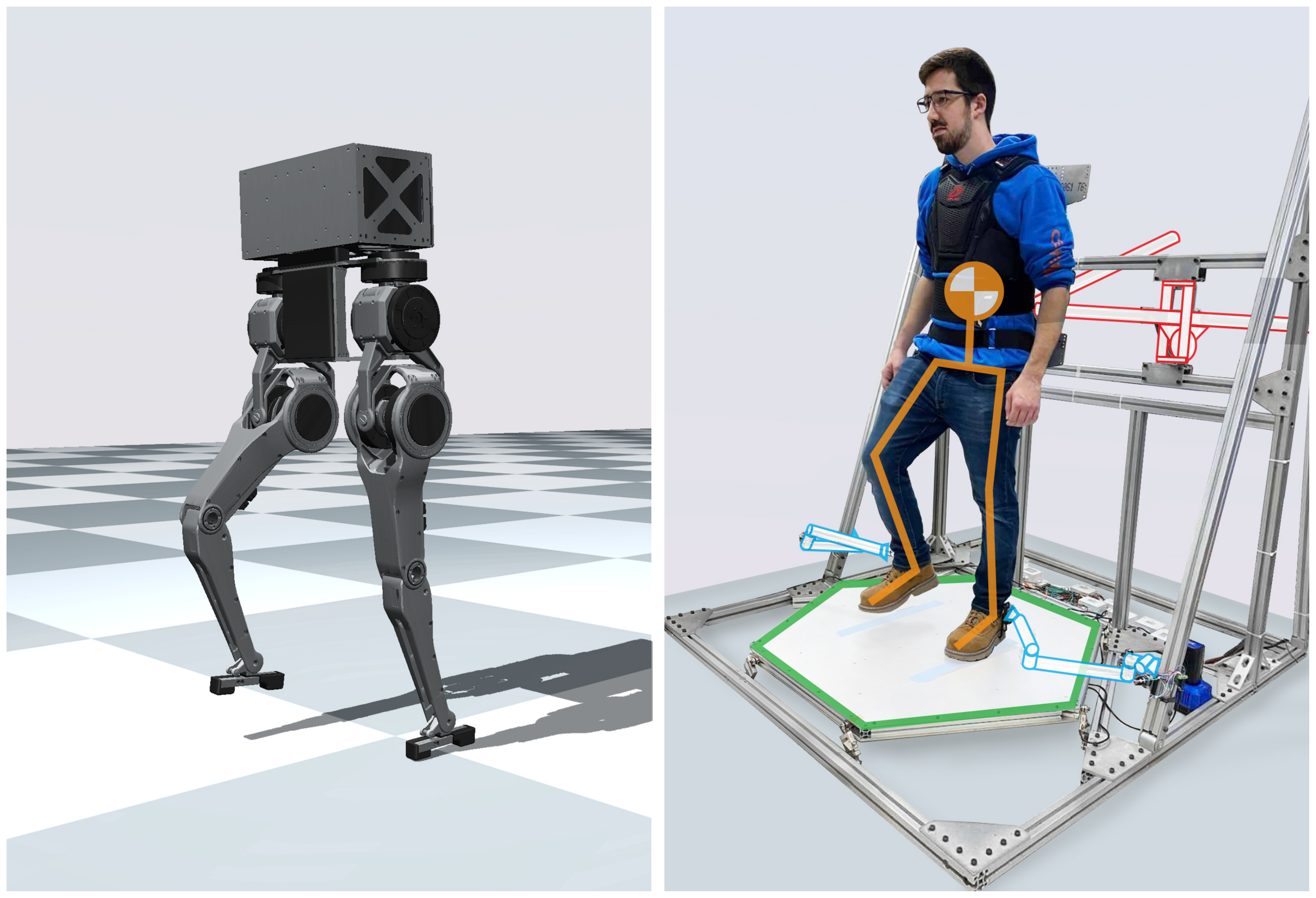}
\end{center}
\caption{The human pilot (right) uses our human-machine interface to dynamically synchronize with the simulated bipedal robot Tello (left) using our whole-body dynamic telelocomotion framework. Supporting video link: \url{https://youtu.be/_ecsf-QxVYg} }
\label{mainfig}
\end{figure}

Synchronizing the gait in real-time between a human and a bipedal robot poses considerable challenges. First, while humans and humanoid robots have a similar morphology, they have fundamental differences. This motivates the need to transfer locomotion strategies from humans to robots. One approach is to abstract human locomotion through reduced-order models \cite{abstractloco_1, Prof_TRO}. The second challenge is that the human and robot have walking dynamics that are destined to evolve at different spatial and temporal scales. This is primarily due to the underactuated nature of walking coupled with the difference in natural frequencies between human and robot. This motivates the use of bilateral feedback between human and robot \cite{Prof_TRO, haptic_fdb_ex2, haptic_fdb_ex3}. In order to inform the human pilot of robot dynamics, haptic force feedback can  encode information about the robot's sense of balance, i.e., stability, in an instinctual manner to harness human sensorimotor skills. The final challenge is the development of specialized hardware. This is why 
human-machine interfaces that facilitate whole-body motion capture and haptic force feedback have been instrumental in advancing teleoperated humanoid robot research \cite{hmi_ex1, hmi_ex2, hmi_ex3}.

Encouraging results have been obtained from the unification of three concepts: abstraction of humanoid locomotion, scaling of human whole-body motion to robot proportions, and dynamic synchronization of human and robot via bilateral feedback. This framework, termed whole-body dynamic telelocomotion, has demonstrated synchronized balancing and stepping between a human pilot and bipedal robot in hardware \cite{SyncWalk_ex1}. In our previous work, it yielded synchronized dynamic walking of a simulated bipedal robot \cite{colin2022bipedal}. The extension from stepping to walking was enabled by a virtual interface linking human and bipedal robot. This interface serves as a virtual reference of human walking behavior generated from their stepping motion. Although similar methods have been proposed, \cite{walk_ref_ex1, walk_ref_ex2}, they lacked integration into a holistic framework to enable the transfer of intelligence and sensorimotor skills from human to robot. 


While the results of \cite{colin2022bipedal} were promising, sustained synchronized locomotion between human and robot was challenging. Consistently reaching more than 1.0~m of robot walking through teleoperation was difficult. The human pilot could generate a walking gait through a step, but struggled to generate transitions between walking gaits. Continuously adapting to robot walking dynamics proved taxing over time. On the other hand, bipedal robot walking using feedback control methods has become highly robust \cite{fdb_ex1, fdb_ex3}. These control methods benefit from considering the hybrid and underactuated dynamics of walking, allowing seamless transitions between walking gaits, particularly at different speeds.

In this work, the hybrid linear inverted pendulum model (H-LIPM) \cite{s2s_hlip_mainref} is utilized to abstract our human walking reference and thus capture the hybrid and underactuated dynamics of walking within our whole-body dynamic telelocomotion framework. The same mapping of human stepping motion to a desired walking gait is utilized as in \cite{colin2022bipedal} to infer human locomotion strategy. This allows the human pilot to intuitively update the H-LIPM dynamics during telelocomotion. Additionally, a reactive robot controller is designed to reproduce scaled human motion on the robot's full-body dynamics. The main contribution of this work lies in incorporating the step-to-step dynamics of walking into our human walking reference interface. The result is dynamically consistent human walking reference motion. Another contribution is the integration of a reactive robot controller into our telelocomotion framework. As depicted in Fig.~\ref{mainfig}, the presented methods are validated through robot full-body dynamics simulation experiments conducted with our telelocomotion framework. The results demonstrate that a human pilot can dynamically synchronize with the humanoid robot Tello to achieve sustained locomotion. This is demonstrated by the pilot's capacity to control robot walking speeds within the range of 0.0~m/s to 0.3~m/s, and maintain robot backward walking for distances of up to 2.0~m.


\section{Human Walking Reference Generation}
\label{hwrmsec}

\begin{figure}[t]
\begin{center}
\includegraphics[width=0.9\linewidth, clip, trim=0mm 140mm 247mm 0mm]{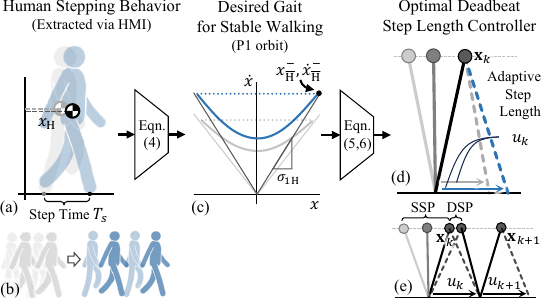}
\end{center}
\caption{The human pilot (a) transitions between desired walking gaits (b) by stepping along the human-machine interface. These gaits are generated using a mapping that translates human stepping into a period 1 orbit. To achieve the target walking behavior, a deadbeat controller is utilized on the human walking reference dynamics, abstracted using the H-LIPM reduced-order model (d). This approach embeds the step-to-step dynamics of walking (e) into our human walking reference motion generation process.}
\label{s2skeyconceptfig}
\end{figure}

The goal of incorporating step-to-step dynamics into our human walking reference (HWR) interface is to facilitate dynamically consistent HWR motion generation by the human pilot. The proposed methodology is presented in this section and summarized in Fig.~\ref{s2skeyconceptfig}. First, a theoretical basis for creating an abstraction of the HWR using a reduced-order model is presented, namely the hybrid linear inverted pendulum model (H-LIPM). Next, a method from \cite{colin2022bipedal} of mapping a human's stepping motion to their desired walking behavior is reviewed. Finally, we highlight the major contribution from this work and detail the HWR motion generation algorithm. 

\subsection{H-LIP Model}
\label{hlipsubsec}

The H-LIPM \cite{hlip_first}, an extension of the LIPM, is a two-domain hybrid system that encapsulates the hybrid dynamics of point-foot walking. The two domains of walking are a single support phase (SSP) and a double support phase (DSP). The domain-specific continuous dynamics of the H-LIPM are governed by
\begin{equation}
\label{HLIPM_dyn}
\begin{aligned}
    &\textrm{SSP:}&&\ddot{x} = \omega^2 x,
\\
    &\textrm{DSP:}&&\ddot{x} = 0,
\end{aligned}
\end{equation}
where $x$ is the center of mass (CoM) position relative to the stance foot, $\omega = \sqrt{\frac{g}{h}}$ is the natural frequency under the gravity field $g$, and $h$ is the CoM height. The H-LIPM describes the discrete transitions between SSP (S) and DSP (D) through smooth hybrid system reset maps, which are defined as follows:
\begin{subequations}\label{eq:discretemap}
\begin{alignat}{2}
\label{discretemap_S2D}
&\Delta_{S\rightarrow D} : & \quad &
\begin{cases}
    x^{+} = x^{-}  \\
    \dot{x}^{+} = \dot{x}^{-} 
\end{cases} 
\tag{\ref{eq:discretemap}a}
,\\
\label{discretemap_D2S}
&\Delta_{D\rightarrow S} : & \quad &
\begin{cases}
    x^{+} = x^{-} - \ell  \\
    \dot{x}^{+} = \dot{x}^{-} 
\end{cases}  
\tag{\ref{eq:discretemap}b}
,
\end{alignat}
\end{subequations}
where the control input $\ell$ is the step length and the superscripts $+$ and $-$ denote CoM states after and before each discrete state reset, respectively. The closed-form solution of the hybrid system is derived by solving Eqs.~\eqref{HLIPM_dyn} -~\eqref{eq:discretemap}. This solution defines the time evolution of the H-LIPM dynamics, as shown below:
\begin{subequations}
\label{HLIPM_sol}
\begin{alignat}{2}
\label{HLIPM_sol_SSP}
&\textrm{SSP:}\quad
&&\begin{cases}
    x(t) = c_{1}e^{\omega t} + c_{2}e^{-\omega t}  \\
    \dot{x}(t) = \omega(c_{1}e^{\omega t} - c_{2}e^{-\omega t})
\end{cases},
\\
\label{HLIPM_sol_DSP}
&\textrm{DSP:}\quad
&&\begin{cases}
    x(t) = x^{-} + \dot{x}^{-}t  \\
    \dot{x}(t) = \dot{x}^{-} 
\end{cases},
\end{alignat}
\end{subequations}
where time in each phase is denoted by $t \in [0, T_{SSP}]$ or $t \in [0, T_{DSP}]$. $T_{SSP}$ and $T_{DSP}$ are the duration of SSP and DSP, respectively. The constants $c_{1} = \frac{1}{2}(x^{+} + \frac{1}{\omega}\dot{x}^{+})$ and $c_{2} = \frac{1}{2}(x^{+} - \frac{1}{\omega}\dot{x}^{+})$ are constants of integration.


\subsection{Human Stepping Motion Mapping}
\label{mappingsubsec}

A mapping of human stepping motion to desired walking motion was introduced in \cite{colin2022bipedal}. The two assumptions made are that 1) the desired human walking gait corresponds to a stable walking gait of the canonical LIPM and 2) the human pilot's ($\sh$) stepping frequency and CoM position along our human-machine interface (HMI) are sufficient for the pilot to intuitively generate these HWR ($\shbar$) gaits.

From a technical perspective, these two assumptions allow mapping the human pilot's stepping frequency, $T_{s}^{-1}$, and CoM position, $x_{\sh}$, to a stable walking gait. The walking gait chosen to represent nominal human walking is characterized as satisfying $\left({x}^{+}, \dot{x}^{+}\right) = \left(-{x}^{-}, \dot{x}^{-}\right)$ during a single step. This one-step periodic orbit, known as a P1 orbit, can be generated in real-time from $T_{s}$ and $x_{\sh}$. The core idea is to map $x_{\sh} \rightarrow \xi_{x\shbar}^{-}$, i.e., the human CoM position along the HMI equates to some desired pre-impact (end-of-step) value of the walking reference's divergent component of motion (DCM). The DCM, defined $\xi_{x}:=x + \frac{\dot{x}}{\omega}$, is the key state tracked in our telelocomotion framework to synchronize human and robot locomotion as it captures the unstable component of motion of the pendulum dynamics \cite{Prof_TRO}.

It can be shown that $\xi_{x}^{-}$ and $T_{s}$ can completely characterize a P1 orbit given some necessary and sufficient conditions are satisfied. Full details can be found in \cite{colin2022bipedal}. In general, from human stepping motion, the desired pre-impact CoM states of a stable walking gait can be computed as following:
\begin{equation}
\label{humanmotionmapping}    
    \begin{cases}
        x_{\shbar}^{-}(T_{s}, x_{\sh}) = x_{\sh} \slash 
        ( {1 + \frac{\sigma_{1\sh}}{\omega_{\sh}}}
        ), 
        \\
        \dot{x}_{\shbar}^{-}(T_{s}, x_{\sh}) = \omega_{\sh}
        (x_{\sh} - x_{\shbar}^{-} ),
    \end{cases}
\end{equation}
where $\sigma_{1\sh}$ is the orbital slope of the human LIPM, defined
$\sigma_{1\sh}:=\omega_{\sh} \: \coth \Big(\frac{T_{s}}{2}\omega_{\sh}\Big).$ We can extend this mapping to stable walking gaits of the H-LIPM.


\subsection{S2S Dynamics Approach}
\label{s2sdynsubsec}

The key idea is to \textit{abstract the HWR as a H-LIPM with a control policy that captures the desired human pilot's walking behavior while obeying the step-to-step dynamics of walking}. In our telelocomotion framework, this will enhance the robot controller's capability to consistently track the motion of the HWR when scaled to robot proportions. 

The approach in \cite{colin2022bipedal} forced the pilot to learn the robot's dynamics of transitioning between different stable walking gaits. The HWR was assumed to be a LIPM whose dynamics \textit{always} evolved along a P1 orbit. In order to aid the pilot, a haptic virtual spring was implemented to give them an acceleration sensation similar to that of the HWR. This allowed for successful HWR motion generation and dynamic synchronized walking between human and bipedal robot, but it was not sustainable for prolonged periods. 

The non-trivial ability to traverse between these stable walking gaits can be more accurately explained by the step-to-step (S2S) dynamics of bipedal walking \cite{s2s_hlip_mainref}. The S2S dynamics of walking are perfectly highlighted by the H-LIPM. As shown in Eq.~\eqref{HLIPM_sol}, under the H-LIPM approximation, the dynamics of walking are purely passive throughout the continuous domains. From Eq.~\eqref{discretemap_D2S}, the control input, $\ell$, is applied only discretely. This motivates viewing the dynamics of walking at the step level. The result is a discrete linear dynamical system,
\begin{equation}
\label{s2sdyneq}
\mathbf{x}_{k+1} = A\mathbf{x}_{k} + Bu_{k},
\end{equation}
that maps step $k$ pre-impact CoM states, $\mathbf{x}_{k}$, to step $k + 1$ pre-impact CoM states, $\mathbf{x}_{k+1}$, via the step length $u_{k} = \ell$. The state transition matrix, $A$, and control input mapping vector, $B$, are defined by combining Eqs.~\eqref{eq:discretemap} -~\eqref{HLIPM_sol}. Thus, stabilization of $\mathbf{x}$ to a desired $\mathbf{x}^{*}$ in Eq.~\eqref{s2sdyneq} can be enforced with feedback control. The case of particular interest is when $\mathbf{x}^{*}$ corresponds to a P1 orbit. In this case, the deadbeat controller from \cite{s2s_hlip_mainref} can optimally drive $\mathbf{e} := (\mathbf{x} - \mathbf{x^{*}}) \rightarrow \mathbf{0}$ in two steps:
\begin{equation}
\label{deadbeatcontrollereq}
u_{k} = x^{-} + x^{-*} + T_{DSP}\dot{x}^{-} + \frac{\coth (T_{SSP}\omega)}{\omega}(\dot{x}^{-} - \dot{x}^{-*}).
\end{equation}

In this work, Eq.~\eqref{deadbeatcontrollereq} is the control policy of the HWR. The control objective is to achieve the desired human locomotion strategy captured by Eq.~\eqref{humanmotionmapping} on the dynamics of the HWR. The dynamics of the HWR evolve according to Eq.~\eqref{HLIPM_sol}. By modeling the HWR as a H-LIPM with step-to-step dynamics given by Eq.~\eqref{s2sdyneq}, the hybrid and underactuated nature of walking is considered in the motion generation of the HWR as outlined in Algorithm~\ref{alg:myalgorithm}.

\begin{algorithm}[t]
  \caption{HWR Motion Generation}
  \footnotesize
  \label{alg:myalgorithm}
  \begin{algorithmic}[1]
    \Require Initialize ($x_{\shbar}^{+}$, $\dot{x}_{\shbar}^{+}$) \& $t$
    \While{Telelocomotion}
      \If{$SSP\rightarrow DSP$ $\lor$ $DSP\rightarrow SSP$}
        \State Set ($x_{\shbar}^{-}$, $\dot{x}_{\shbar}^{-}$) = ($x_{\shbar}$, $\dot{x}_{\shbar}$)
        \If{$SSP\rightarrow DSP$}
          \State Update ($x_{\shbar}^{+}$, $\dot{x}_{\shbar}^{+}$) with Eq.~\eqref{discretemap_S2D}
        \ElsIf{$DSP\rightarrow SSP$}
          \State Update ($x_{\shbar}^{+}$, $\dot{x}_{\shbar}^{+}$) with Eq.~\eqref{discretemap_D2S}
        \EndIf
        \State Reset $t = 0$
      \Else
        \If{$SSP$}
          \State Update ($x_{\shbar}$, $\dot{x}_{\shbar}$) with Eq.~\eqref{HLIPM_sol_SSP}
          \State Update $\mathbf{x}^{*}$ with Eq.~\eqref{humanmotionmapping}
          \State Predict $\mathbf{x}$ with Eq.~\eqref{HLIPM_sol_SSP}
          \State Compute $u_{k}$ with Eq.~\eqref{deadbeatcontrollereq}
        \ElsIf{$DSP$}
          \State Update ($x_{\shbar}$, $\dot{x}_{\shbar}$) with Eq.~\eqref{HLIPM_sol_DSP}
        \EndIf
      \EndIf
    \EndWhile
  \end{algorithmic}
\end{algorithm}

\section{Robot Controller}
\label{robotcontrolsect}

The other contribution of this work is the integration of a reactive robot controller into our whole-body dynamic telelocomotion framework to facilitate dynamic synchronized locomotion between human pilot and bipedal robot. The objective is to reactively reproduce scaled human motion and implement the human pilot's locomotion strategy. This strategy is abstracted using reduced-order models. As a result, in our telelocomotion framework, dynamic synchronization between human and robot is enforced at the reduced-order model level. The role of the robot controller is to track the resulting robot 2D LIP trajectory at the full-body dynamics level. This trajectory represents stable locomotion (center of mass dynamics) and the corresponding control input (center of pressure strategy). 

\begin{figure}[t]
\begin{center}
\includegraphics[width=0.5\linewidth, clip, trim=1.5mm 118.5mm 263mm 1.25mm]{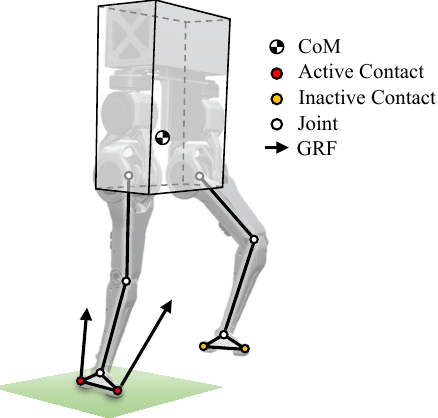}
\end{center}
\caption{The SRBM approximation of the full-body dynamics of the bipedal robot Tello (shown here during $SSP$). The floating-base dynamics of the robot are actuated by ground reaction forces (GRFs) applied at active contact points. These contact points are located at the toe and heel of the robot's feet.}
\label{robotmodels_fig}
\end{figure}

The robot controller considers the full-body dynamics of the humanoid robot Tello \cite{tello_main}. The controller consists of an optimization-based balance controller that computes joint torques for the legs in stance during DSP and SSP. As shown in Fig.~\ref{robotmodels_fig}, the balance controller approximates the robot full-body dynamics using the single rigid-body model (SRBM). The controller also consists of a swing-leg controller that computes joint torques for the leg in swing during SSP. 

\begin{figure*}[t]
\begin{center}
\includegraphics[width=0.75\linewidth, clip, trim=0mm 0mm 0mm 0mm]{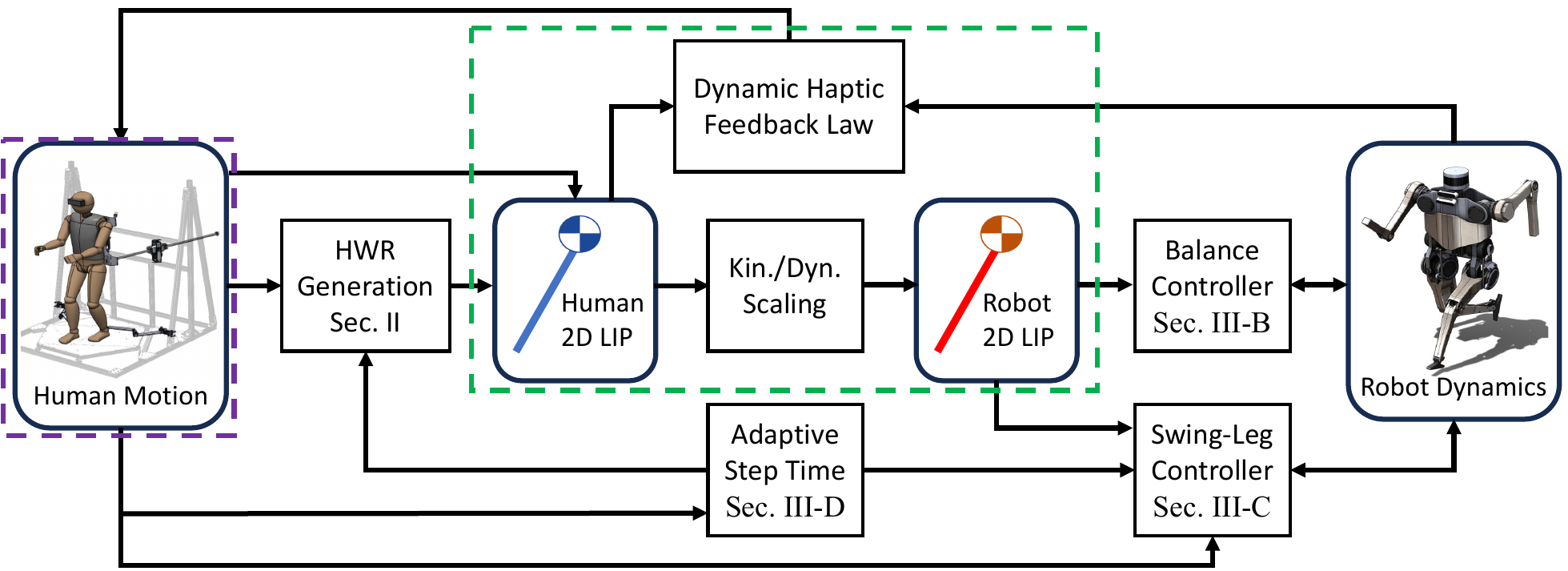}
\end{center}
\caption{High-level overview of our whole-body dynamic telelocomotion framework. The sub-components discussed in this work are labeled with their respective sections for ease of navigation. Previous work is highlighted by the dashed colored lines. For example, the HMI (purple color) in \cite{hmi_ex3} and dynamic synchronization via bilateral feedback teleoperation (green color) in \cite{colin2022bipedal}.}
\label{telelocomotion_framework_fig}
\end{figure*}

\subsection{Robot Model}
\label{robotmodelsubsect}

The 15.8 kg, 16 degree-of-freedom (DoF) bipedal robot Tello has the following full-body dynamics:
\begin{equation}
\label{fullbodydyn}
    M(q)\ddot{q} + C(q, \dot{q})\dot{q} + G(q) = S^{\top}\tau + J_{c}^{\top}f,
\end{equation}
where $q$ is the robot generalized coordinates vector, $M$ is the mass matrix, $C$ is the Coriolis matrix, $G$ is the gravity vector, $S$ is the actuated DoF selection matrix, $J_{c}$ is the contact Jacobian matrix, $\tau$ is the joint torque vector, and $f$ is the ground reaction wrench vector.

A popular approximation of Eq.~\eqref{fullbodydyn} is the single rigid-body model (SRBM) \cite{srbm_1, srbm_3}. The SRBM is a simplified model of legged robots with low centroidal inertia isotropy (CII) \cite{tello_main}, which neglects the inertial effects introduced by the robot's limbs. Thus, the torso's translational and rotational dynamics are only considered. The states of the torso (SRB) are defined as $s:=\left[p, \dot{p}, R, ^{B}\omega \right]^{\top}$, where $p$ is the CoM position vector, $R$ the rotation matrix of the body frame relative to the world frame, and $^{B}\omega$ is the angular velocity vector expressed in body frame, and $\{ B \}$ is the body frame. The SRBM state-space dynamics are the following:
\begin{equation}
\label{SRBM_dyn}
    \dot{s} = 
    \begin{bmatrix} 
        \dot{p} \\ \ddot{p} \\ \dot{R} \\ {}^{B}\dot{\omega} 
    \end{bmatrix} =
    \begin{bmatrix} 
        \dot{p} \\ \frac{1}{m}F + a_{g} \\ R \cdot {}^{B}\omega \\ {}^{B}I^{-1}[R^{\top}\Tilde{\tau} - {}^{B}\hat{\omega} {}^{B}I {}^{B}\omega]
    \end{bmatrix}     
\end{equation}  
where $m$ is the robot mass, $^{B}I$ is the torso inertia matrix in the body frame, $a_{g}$ is the gravity vector, $F$ and $\Tilde{\tau}$ are translational and rotational components of the net wrench acting on the CoM, respectively, and $\hat{\left( \cdot \right)}$ is the operator that maps an element of $\mathbb{R}^3$ to the 3D rotation group $SO(3)$. 
The net wrench, $\mathcal{F}$, arises from the ground reaction forces (GRFs) generated at active contact points, i.e., 
\begin{equation}
\label{netwrench_eq}
    \mathcal{F} = 
    \begin{bmatrix}
        F \\ \Tilde{\tau}
    \end{bmatrix} = \sum_{i=1}^{n} 
    \begin{bmatrix}
        \mathbb{I} \\ \hat{r}_{i}
    \end{bmatrix}\Tilde{f}_{i},
\end{equation}
where $n$ is the number of active contact points, $\mathbb{I}$ is the identity matrix, $r_i$ represents the vector from the CoM to the $i$-th active contact point, while $\Tilde{f}_i$ denotes the GRF vector at that contact point. For Tello, which has line feet, we assume two contact points at the toe and heel of each foot \cite{srbm_3}.

\subsection{Optimization-Based Balance Controller}
\label{balancecontrollersubsect}

The balance controller computes the GRF vectors required to reactively track a desired SRBM trajectory. A PD task-space controller is implemented, 
\begin{equation}
\label{torsoPDeq}
    \mathcal{F}^{d} = K_{p}(q_{s}^{d} - q_{s}) + K_{d}(\dot{q}_{s}^{d} - \dot{q}_{s}),
\end{equation}
to compute the desired net wrench, $\mathcal{F}^{d}$, to track the desired task-space trajectory, $q_{s}^{d}$ and $\dot{q}_{s}^{d}$. The task-space coordinate vector, $q_{s}$, of the SRB is defined as $q_{s}:=[p, \Theta]^{\top}$, where $\Theta$ is the euler-angle representation of $R$. Appropriate coordinate transformations are made to construct $q_{s}$ and $\dot{q}_{s}$ from $s$. $K_{p}$ and $K_{d}$ are diagonal proportional and derivative gain matrices, respectively.

Once $\mathcal{F}^{d}$ is determined using Eq.~\eqref{torsoPDeq}, a force distribution problem is formulated as a quadratic program (QP). This QP aims to calculate the corresponding GRF vectors at the active contact points, satisfying Eq.~\eqref{netwrench_eq}. Furthermore, the GRF vectors undergo projections from task-space to both the friction cone and the motor-space, where linear inequality actuation constraints are enforced. This accounts for both the parallel actuation scheme of Tello \cite{tello_main} and the effects of friction. GRF vectors are also constrained to not exert a pulling force on the ground. Thus, the actuation-aware force distribution QP takes the following form:
\begin{equation}
\label{forcedist_eq}
\begin{aligned}
\min_{\Tilde{f}} \quad & w_{1}\| \mathcal{F} - \mathcal{F}^{d} \|_{2}^{2} + w_{2}\| \Tilde{f} - \Tilde{f}_{0} \|_{2}^{2} \\
\text{s.t.} \quad & |J_{MJ}^{\top}J_{JT}^{\top}R_{BW}\Tilde{f}| \leq \tau_{m}(q_{j}, \dot{q}_{j}) \quad \text{(motor lim.)} \\
& |\Tilde{f}_{x,y}| \leq \mu \cdot \Tilde{f}_{z} \quad \text{(friction cone)}\\
& \Tilde{f}_{z} \geq 0 \quad \text{(no-pull)}
\end{aligned}
\end{equation}
where $\Tilde{f}$ is the collection of GRF vectors at active contact points, $\Tilde{f}_{0}$ is the previous value of $\Tilde{f}$, $w_{1}$ and $w_{2}$ are scalar weighting factors, $R_{BW}$ is a matrix of rotation matrices $R^{\top}$ that map $\Tilde{f}$ from the world frame to the body frame, $J_{JT}$ is a matrix of contact Jacobians that map task-space forces to joint torques, $J_{MJ}$ is a matrix of topology Jacobians that map joint torques to motor torques, $\tau_{m}$ is a vector of configuration and speed-dependent motor torque limits, $q_{j}$ and $\dot{q}_{j}$ are leg joint positions and velocities, respectively, and $\mu$ is the friction coefficient. The second term of the cost function in Eq.~\eqref{forcedist_eq} acts as a solution filter. 

Finally, joint torques for legs in stance are set using the SRBM approximation of Eq.~\eqref{fullbodydyn}. However, to maintain the task-space configuration of the robot feet, joint-space posture control is appended to the optimization-based balance controller from Eq.~\eqref{forcedist_eq}. Consequently, the final expression for the joint torques of legs in stance, $\tau_{st}$, is the following:
\begin{equation}
\label{stlegtq_eq}
    \tau_{st} = J_{JT}^{\top}R_{BW}\Tilde{f} + \tau_{po},
\end{equation}
where $\tau_{po}$ is vector of posture control joint torques. A sigmoid function is utilized to smoothly transition between stance-leg joint torques and swing-leg joint torques. The computation of $\tau$ from Eq.~\eqref{fullbodydyn} is the summation of smoothed stance-leg and swing-leg joint torques.

\subsection{Swing-Leg Controller}
\label{swinglegcontrollersubsect}

During SSP, when one of the robot's legs is in swing, corresponding joint torques cannot be computed using Eq.~\eqref{forcedist_eq}. This is because the leg is no longer in contact with the ground and as a result a GRF cannot be generated. Instead, the goal of the swing-leg controller is to map the robot's foot position at the start of SSP to a location by the end of SSP that achieves the desired step placement. 

This mapping of the robot's foot during SSP is accomplished via smooth task-space trajectories similar to in \cite{fdb_ex3}. On the transverse plane, specifically the $x$ or $y$ axis, the function,
\begin{equation}
\label{sw_traj_xy_eq}
    p^{x,y}_{sw}(s) = \frac{1}{2} \biggl( (1 + \cos(\pi s))p^{i}_{sw} + (1 - \cos(\pi s))p^{f}_{sw} \biggr),
\end{equation}
maps the robot's initial foot position, $p^{i}_{sw}$, to a final desired foot position, $p^{f}_{sw}$, along each axis. The phase variable, denoted as $s \in [0, 1]$, represents the normalized progression of time during a step. On the other hand, $z$-position trajectories are defined differently. The objective is to re-establish contact with the ground at time $s = 1$. The following function, 
\begin{equation}
\label{sw_traj_z_eq}
    p^{z}_{sw}(s) = p^{i}_{sw} + \frac{z_{cl}}{2} \biggl[1 - \cos(2\pi s) \biggr],
\end{equation}
characterized by a maximum step height, $z_{cl}$, attains the intended foot z-position motion.

In this work, swing-leg joint torques, $\tau_{sw}$, are computed using joint-space PD control. First, utilizing inverse kinematics, Eqs.~\eqref{sw_traj_xy_eq} -~\eqref{sw_traj_z_eq}, along with their derivatives, are transformed into desired joint positions, $q^{d}_{j}$, and corresponding desired joint velocities, $\dot{q}^{d}_{j}$, trajectories. Then, the joint-space trajectories are tracked using the PD control law,
\begin{equation}
\label{sw_leg_control_eq}
    \tau_{sw} = K^{sw}_{p}(q^{d}_{j} - q_{j}) + K^{sw}_{d}(\dot{q}^{d}_{j} - \dot{q}_{j}),
\end{equation}
where $K^{sw}_{p}$ and $K^{sw}_{d}$ are diagonal proportional and derivative gain matrices, respectively.

\subsection{Integration with Whole-Body Dynamic Telelocomotion Framework}
\label{integrationtelelocomotionsubsect}

The robot controller, composed of the optimization-based balance controller and swing-leg controller, is integrated into our whole-body dynamic telelocomotion framework (Fig.~\ref{telelocomotion_framework_fig}). Although the robot controller is capable of operating independently, it was designed to be purely reactive, with locomotion planning being entrusted to the human pilot.

The human locomotion strategy is represented by a 2D LIP trajectory. This trajectory is constructed using dynamic human motion data in the frontal plane, combined with the human walking reference generated from human stepping motion data along the sagittal plane. The trajectory is then scaled to robot proportions. The goal of the robot controller is to reproduce the resulting robot 2D LIP trajectory at the full-body (3D) dynamics level. 

The reproduction of the robot's 2D LIP trajectory is described next. The $x$-$y$ components of the desired robot CoM wrench from Eq.~\eqref{forcedist_eq} are given by robot forces that track the desired scaled human motion \cite{Prof_TRO}. Additionally, the LIPM assumptions of constant CoM height and the omission of angular dynamics are enforced by Eq.~\eqref{torsoPDeq}. The $x$-$y$ swing-leg trajectory from Eq.~\eqref{sw_traj_xy_eq} is updated to keep dynamic synchronization between the human pilot and the robot at step transitions. Specifically, in the $x$-direction, the final desired robot foot position maintains tracking of the scaled HWR at the commencement of the next step. In the $y$-direction, the final desired robot foot position maintains equal scaled feet width between the human and robot. 

The main challenge in tracking the desired robot 2D LIP trajectory is that the single support time $T_{SSP}$ is not known a priori. $T_{SSP}$ is normally fixed in bipedal walking controllers. Thus, to resolve this issue we leverage the consistency of human motion to implement a data-driven adaptive step timing strategy. In particular, we assume human swing-foot z-position motion approximately follows Eq.~\eqref{sw_traj_z_eq}, parameterized by $s$ (normalized SSP time) and $z_{cl}$. During each step, we employ a damped least-squares method to continuously fit human swing-foot z-position data to Eq.~\eqref{sw_traj_z_eq}. This allows the pilot to control the robot's step height and vary the robot's stepping frequency. This ensures the pilot and the robot will touch-down synchronously. The updated estimate of SSP duration is used to update the HWR and the $x$-$y$ swing-leg trajectories for more robust telelocomotion.

\section{Simulation Results \& Discussion}
\label{resultsexpsec}

\begin{figure}[t]
\begin{center}
\includegraphics[width=0.85\linewidth]{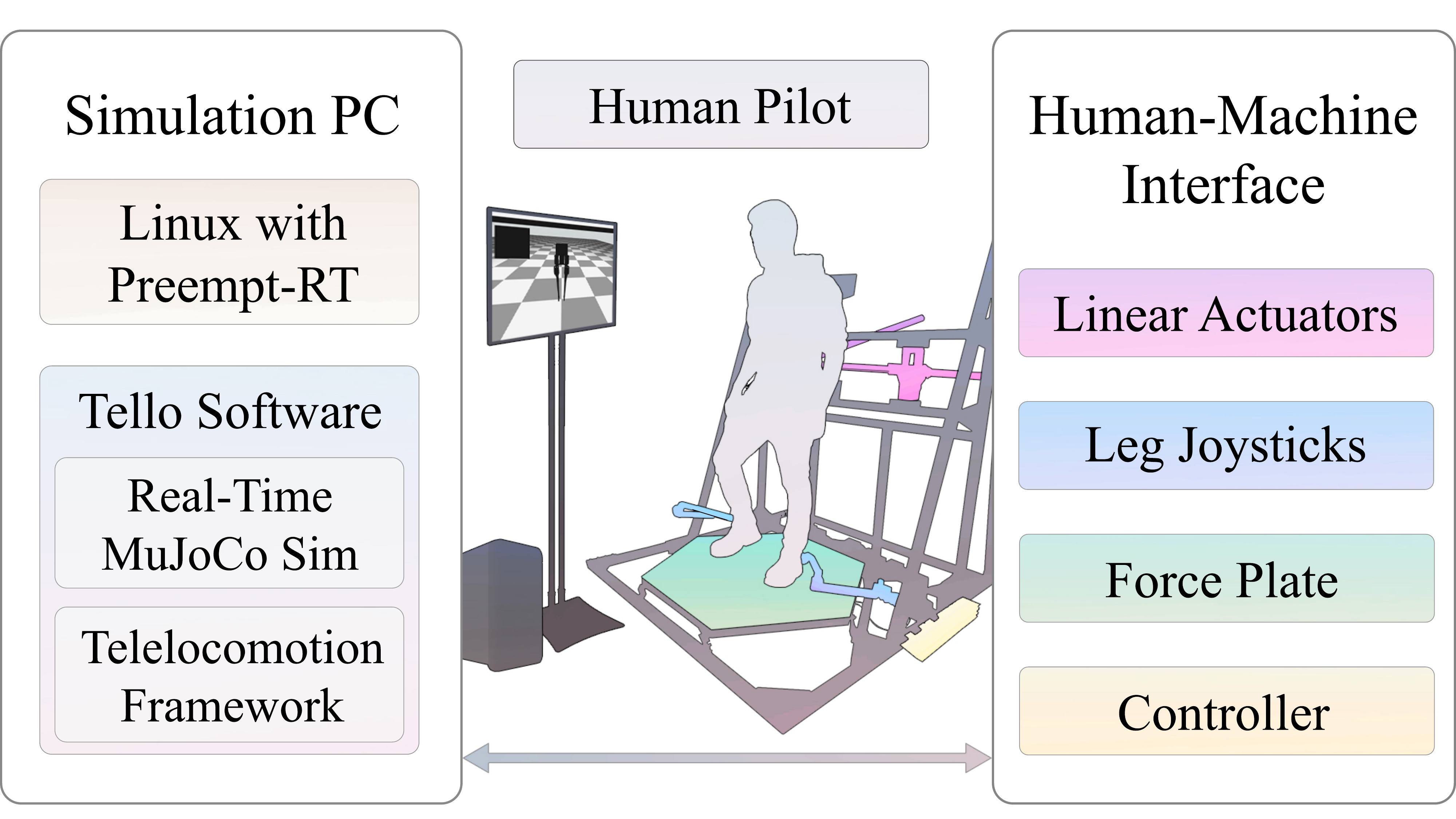}
\end{center}
\caption{Setup of telelocomotion simulation experiments. Real-time telelocomotion of the simulated robot by the human pilot is enabled by UDP communication between our HMI and simulation PC.}
\label{exp_setup_fig}
\end{figure}

To validate our approach to human walking generation and assess the performance of our robot controller in executing the human's locomotion strategy, we designed simulation telelocomotion experiments. As shown in Fig.~\ref{exp_setup_fig}, during these experiments, our whole-body dynamic telelocomotion framework is implemented in a hardware-simulation set-up. As depicted in Fig.~\ref{mainfig}, the hardware, our human-machine-interface (HMI), is used by a human pilot to control the dynamics of the bipedal robot Tello in a simulation environment with the MuJoCo physics engine \cite{todorov2012mujoco}. 

The first experiment involves velocity tracking, where the human pilot tracks a moving target at increasing speeds (0.1~m/s, 0.2~m/s, and 0.3~m/s) for a total of 6.0~m of synchronized robot walking. A square target is displayed on the ground, and the objective is to generate a walking reference that transitions smoothly between stable walking gaits at different speeds and stepping frequencies. This experiment assesses both the human pilot's ability to generate a walking reference using Algorithm~\ref{alg:myalgorithm} and the robot controller's capability to execute the desired human locomotion strategy. The second experiment focuses on synchronized backward walking between the human and robot. It demonstrates similar capabilities as the first experiment but highlights an important telelocomotion behavior for effective navigation in an environment.

In this section, we outline the experimental setup employed in our telelocomotion experiments. The focus is on describing the connection between our HMI and a simulation PC, enabling real-time telelocomotion of a simulated bipedal robot. Subsequently, we showcase the successful completion of experiments 1 \& 2 by a trained human pilot, and proceed to analyze the results in the context of the presented methods. Finally, we discuss some limitations of our approach. 

\subsection{Experimental Setup}
\label{expsetupsubsect}

During telelocomotion simulation experiments, a trained human pilot utilizes our HMI to dynamically teleoperate the simulated bipedal robot Tello, whose dynamics and corresponding controller are running in the simulation PC in real-time. Communication between the HMI and the simulation PC is established via UDP. As shown in Fig.~\ref{exp_setup_fig}, the human pilot receives visual feedback through a computer monitor, which displays the rendered simulation. This is critical in both experiments as visual feedback is required for velocity target tracking and walking backwards a specific distance. 

As illustrated in Fig.~\ref{exp_setup_fig}, our HMI consists of linear actuators, lower-body leg joysticks, a force plate, and a real-time embedded controller. The linear actuators perform dual functions of measuring the human CoM while simultaneously applying haptic forces to the human's torso. The lower-body leg joysticks capture and measure spatial information pertaining to the locations of the human feet. The force plate measures the net contact wrench applied by the human. The actuators are controlled and the motion capture data is collected by a cRIO-9082 (National Instruments) computer.

The simulation of the full-body dynamics of the bipedal robot Tello, described by Eq.~\eqref{fullbodydyn}, is executed using the physics engine MuJoCo. The simulation is seamlessly integrated into a unified program alongside the Tello control software. This software is parallelized across nine distinct CPUs, with all threads running at a frequency of 1~kHz. 

\begin{figure}[t]
\begin{center}
\includegraphics[width=0.99\linewidth]{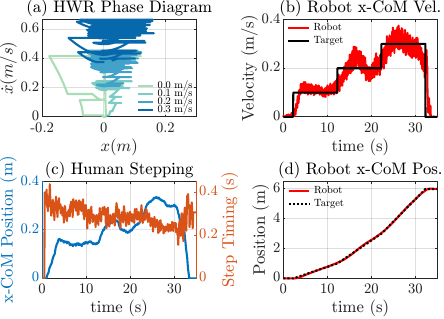}
\end{center}
\caption{Results of velocity tracking experiment I. (a) color-coded HWR phase diagram by robot target velocity, (b) robot CoM x-velocity vs. target, (c) human stepping motion, (d) robot CoM x-position vs. target.}
\label{exp1_results_fig1}
\end{figure}

\subsection{Results \& Discussion}
\label{resultsdiscsubsec}

\begin{figure}[t]
\begin{center}
\includegraphics[width=0.96\linewidth, clip, trim=0mm 0mm 0mm 0mm]{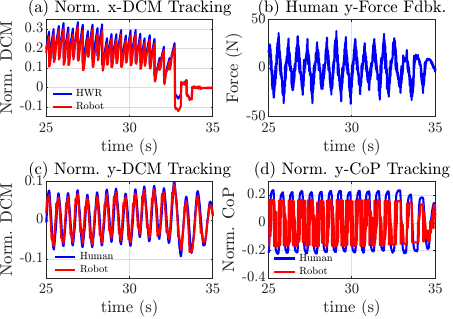}
\end{center}
\caption{Results of velocity tracking experiment II. Normalized DCM tracking performance in (a) sagittal plane and (b) frontal plane, plus corresponding haptic force feedback to human (c) and (d) normalized CoP tracking during final 10.0~s of the experiment.}
\label{exp1_results_fig2}
\end{figure}

Using our framework, depicted in Fig.~\ref{telelocomotion_framework_fig}, along with visual feedback, a trained human pilot successfully completed both experiments. These experiments serve as a compelling demonstration, showcasing the synchronized locomotion between a human and a bipedal robot in tasks involving velocity target tracking and backward walking. The completion of these tasks show the effectiveness of our HWR motion generation method and our reactive robot controller.

The results of the velocity target experiment are shown in Figs.~\ref{exp1_results_fig1} -~\ref{exp1_results_fig2}. Here, the human pilot controls the robot's locomotion to ensure tracking of the target, as illustrated in Fig.~\ref{exp1_results_fig1}(b). Notably, the pilot achieves 6.0 m of synchronized locomotion with the robot, as seen in Fig.~\ref{exp1_results_fig1}(d), affirming the viability of sustained telelocomotion using our methods. Importantly, the pilot regulates the robot's CoM velocity around the target velocity. The key lies in intuitively capturing the human pilot's locomotion strategy from their stepping motion (Fig.~\ref{exp1_results_fig1}(c)). Subsequently, the stepping motion is mapped to a desired walking gait, which serves as the basis for the HWR motion in Fig.~\ref{exp1_results_fig1}(a). Indeed, as seen in Fig.~\ref{exp1_results_fig1}(a), the abstraction of the HWR as a H-LIPM enables generating dynamically consistent walking references, allowing for smooth transitions between desired walking gaits by considering the step-to-step dynamics of walking. Lastly, the ability of the robot controller to reactively reproduce the desired robot 2D LIP trajectory is highlighted by the normalized DCM tracking performance along the sagittal plane (Fig.~\ref{exp1_results_fig2}(a)) and along the frontal plane (Fig.~\ref{exp1_results_fig2}(b)). As seen in Fig.~\ref{exp1_results_fig2}(c), this leads to consistent and bounded haptic force feedback to the human pilot along the frontal plane. This feedback assists the human pilot in maintaining stepping synchronization with the robot. As discussed in our previous work \cite{colin2022bipedal, Prof_TRO}, this is evident from the normalized center of pressure (CoP) synchronization between the two systems, as shown in Fig.~\ref{exp1_results_fig2}(d).

\begin{figure}[t]
\begin{center}
\includegraphics[width=0.98\linewidth, clip, trim=0mm 0mm 0mm 0mm]{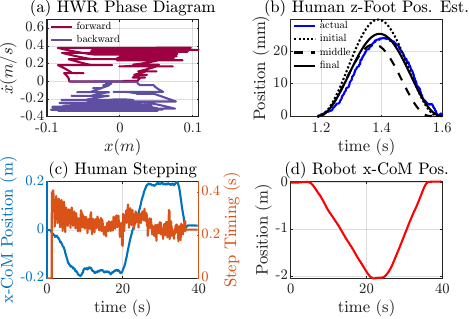}
\end{center}
\caption{Results of backward walking experiment. (a) color-coded HWR phase diagram by walking type, (b) human swing-foot z-position prediction for a single step, (c) human stepping motion, (d) robot CoM x-position.}
\label{exp2_results_fig}
\end{figure}

The results of the backward walking experiment are shown in Fig.~\ref{exp2_results_fig}. The human pilot dynamically synchronizes their backward stepping with bipedal robot backward walking, covering a distance of 2.0~m (Fig.~\ref{exp2_results_fig}(d)), before proceeding to walk forward to the initial position. Like in the first experiment, the human pilot's stepping motion is utilized to generate a HWR motion trajectory, as seen in Fig.~\ref{exp2_results_fig}(a) and Fig.~\ref{exp2_results_fig}(c). However, by walking backwards, the pilot commands desired walking gaits characterized by negative orbital energy, as shown in Fig.~\ref{exp2_results_fig}(c). Additionally, to demonstrate the efficacy of our adaptive step time prediction approach, we analyze the first step from the human stepping data. As shown in Fig.~\ref{exp2_results_fig}(b), we continuously monitor the human z-foot position data and fit it to the presumed swing-leg z-trajectory curve. Thus, allowing us to predict both the duration ($T_{SSP}$) and the height of the step ($z_{cl}$). The prediction accuracy improves with the accumulation of more data, closely approximating the actual human z-foot position throughout the entire step, particularly towards the end. This accurate prediction ensures the dynamic synchronization between the human and bipedal robot at impact events and gives the human the capability to regulate the frequency of the synchronized stepping between the two bipedal systems.

The obtained results show promise in achieving robust telelocomotion of humanoid robots, enhancing their potential as physical avatars. These results also represent tangible progress from the work in \cite{colin2022bipedal}, specifically in successfully realizing sustainable synchronization between human and bipedal robot locomotion in a full-body dynamics simulation. Previously, in a reduced-order dynamics simulation, only 1.28~m of forward walking (0.36~m/s max speed) was achieved. However, it is important to acknowledge the limitations inherent to this work. First, the walking speed of the robot is currently still limited to approximately 0.35~m/s, which, when considering a robot like Tello, corresponds to a human walking speed of 0.5~m/s. While it is possible for the robot to reach faster walking speeds, it becomes challenging for the pilot to sustain telelocomotion at these speeds. The most common failure is the robot lagging behind the HWR due to prolonged periods of DSP by the human that increases the error in normalized DCM between the two systems at the commencement of the next step. Second, it is essential to emphasize that significant training is required for a pilot to reach the demonstrated level of performance. For instance, in the velocity target tracking experiment, the pilot faced difficulties in abruptly stopping the robot from a walking speed of 0.3~m/s. Similarly, transitioning from backward walking to forward walking in the second experiment proved equally challenging. Lastly, it is worth noting that the telelocomotion behaviors enabled by our framework are presently constrained due to the utilization of linear pendulum models to abstract the locomotion of both the human and bipedal robot. To attain a broader spectrum of behaviors, including loco-manipulation tasks, it will be necessary to employ more dynamically rich reduced-order models that can fully exploit the human's whole-body dynamics strategy. 
\section{Conclusion}
\label{conclusionsect}

This paper presents an approach that enhances dynamically consistent walking trajectories within a comprehensive whole-body dynamic telelocomotion framework. By integrating the hybrid and underactuated characteristics of walking, we effectively capture the locomotion strategy of a human pilot through their stepping behavior. A reactive robot controller enables telelocomotion of the full-body dynamics of a bipedal robot. Real-time telelocomotion simulation experiments demonstrate the synchronization between a trained human pilot and a bipedal robot, including tracking a moving target and synchronized backward walking.

Future work will focus on implementing these methods on the real humanoid robot Tello, enabling dynamic control for navigating 3D environments. We will develop a human motion mapping for robot turning and address performance limitations related to walking speed. Additionally, we aim to incorporate human predictive intent into our human walking interface, leveraging human intelligence to a greater extent.




\bibliographystyle{unsrt}
\bibliography{main.bib}

\end{document}